\documentclass{article}

\usepackage[preprint]{neurips_2020}

\usepackage[utf8]{inputenc} 
\usepackage[T1]{fontenc}    
\usepackage{hyperref}       
\usepackage{url}            
\usepackage{booktabs}       
\usepackage{amsfonts}       
\usepackage{nicefrac}       
\usepackage{microtype}      
\usepackage{setspace}
\usepackage{tcolorbox}
\definecolor{mygreen}{RGB}{12, 145, 0}
\definecolor{bittersweet}{rgb}{1.0, 0.44, 0.37}
\usepackage[title]{appendix}

\title{Automatic Story Generation:\\ Challenges and Attempts}

\author{%
  Amal Alabdulkarim \thanks{Equal contributions} , Siyan Li \footnotemark[1] , Xiangyu Peng \footnotemark[1] \\
  Georgia Institute of Technology\\
  Atlanta, GA 30332 \\
  \texttt{\{amal, lisiyansylvia, xpeng62\}@gatech.edu}
}

\begin{document}

\maketitle



\section{Introduction and Motivation}

Storytelling is central to human communication. People use stories to communicate effectively with one another. As humans, we engage with well-told stories and comprehend more information from stories \cite{suzuki_dialogues_2018}. However, when it comes to automatic storytelling, computers still have a long way to go. The field of automated story generation, or computational narrative, has received more attention because of recent technological enhancements. The importance of computational narrative is that it can improve human interaction with intelligent systems. Storytelling helps computers communicate with humans \cite{riedl_computational_2016}, and automated story generation drives improvements in natural language processing. Computational narrative research involves story understanding, story representation, and story generation. In this survey, we will focus on the story generation capabilities of computational systems. 

Many surveys were written on different facets of computational storytelling. \cite{gervas2009computational} provides a chronological summary of storytelling systems focusing on computational creativity, measured using metrics including the stories' novelty and the users' involvement in the storytelling process. \cite{riedl2013interactive} focuses on interactive intelligence, a digital interactive storytelling experience where users interact with the computational system to build storylines. The survey paper touches on generating narrative structures and character building. \cite{riedl_computational_2016} discusses human-centered computational narrative and how it can improve artificial intelligence applications. The paper shed some light on machine learning challenges concerned with story generation and commonsense reasoning. Nevertheless, it does not go into these challenges in-depth as it is not its primary focus point. 

Past survey papers focused primarily on story generation using specific approaches or on specific sub-problems in story generation. For example, \cite{7439785} summarizes progress in the areas of plot and space generation without much discussion around neural language models. \cite{survey_deep} examine different deep learning models used in story generation and categorize them by their goals. However, there is still motivation to organize a survey in a different manner. The process of automatically generating a logically-coherent and interesting narrative is complex. Therefore, it might be more beneficial detailing the major problems present in the field and techniques used to address them rather than summarizing different types of models. For people who are new in the field, our survey should serve as a decent starting point for conducting innovative research in the field. 

Some of the survey papers, albeit comprehensive, do not include the latest development in story generation because of transformers. \cite{riedl2013interactive} chronicles interactive narrative prior to 2013, yet the discussed approaches do not include large-scale neural language models, which we have access to now and has been fueling new research in the field. Another example would be the paper by \cite{gervas2009computational}, where the author comments on storytelling systems and different evaluation criteria for creativity; similarly, all of the systems consist of planning and no neural approaches. 

We acknowledge that more survey papers exist with different areas of focus within the domain of computational narratives, such as Narrative theories \citep{cavazza_narratology_2006}, Interactive Intelligence \citep{luo_review_2015}, Drama Management \citep{roberts_survey_2008}, Plan-based story generation \citep{young_plans_2013}. 

It has been demonstrated that the field of automated story generation has a gap in up-to-date survey papers. Our paper, by laying out all the prominent research problems in story generation and previous literature addressing these issues, will fill this gap. 

The scope of this survey paper is to explore the challenges in automatic story generation. We hope to contribute in the following ways: 
\begin{enumerate}
    \item Explore how previous research in story generation addressed those challenges. 
    \item Discuss future research directions and new technologies that may aid more advancements. 
    \item Shed light on emerging and often overlooked challenges such as creativity and discourse. 
\end{enumerate}

There are several important background concepts crucial to understanding the problem of story generation. Automated story generation is a process involving the use of computer systems to create written stories, often involving artificial intelligence (AI). Story generation requires story understanding and representation, which are usually handled by natural language processing. Hence, the first concentration in this paper is content encoding and comprehension. A system is conventionally defined as capable of story comprehension if it, given a textual story, can read and answer questions about it \citep{lehnert1983boris, reeves1991computational}. Recently, state-of-the-art neural text generation models (such as GPT-2 \citep{radford2019language}), are used to generate stories. These models are trained on the WebText corpus, a collection of texts scraped from the internet. Hence, the key challenge of applying these language models to story generation is to ensure that the generated story remains on topic and maintains entity and event consistencies. In our paper, we consider the following two concepts as crucial starting points: Controllability -- having human inputs influence the generation results (Section~\ref{sec:co} of the paper), and commonsense -- narrative systems with pre-existing knowledge that would help generate coherent stories (Section~\ref{sec:cs} of the paper).

\section{Method}
\subsection{Controllability in Story Generation}
\label{sec:co}
The controllability problem in story generation is the user input's ability to influence the generation results. Such influence often takes the form of a plot the user wishes the system to adhere to when producing a new narrative. Controlling story generation is a significant challenge that gained more attention in the last few years due to the limitations of neural-based story generation approaches. Most modern story generators use Neural based techniques that need little to no manual modeling to generate stories. Neural based models solve the lack of novelty issues found in the symbolic systems due to their unstructured generation. Yet, this advance comes at the cost of less controllability and plot coherence. In this section, we shed light on a few approaches to the problem of controllability, discuss their strengths and weaknesses, and compare their methodologies. 

\textbf{Reinforcement Learning}. \cite{tambwekar2018controllable} aimed at controlling the story plot by controlling its ending and events order. They proposed a deep reinforce approach to controlled story generation with a reward shaping technique to optimize the pre-trained sequence to sequence model in \citep{martin2017event}. Their reward function encompasses two main parts, the distance to the goal verb and the story verb frequency. The distance to the goal verb measures how many lines between a generated verb and the goal verb in training stories. Simultaneously, the story verb frequency counts the stories with both the goal verb and the generated verb. They evaluated their model on plot coherence and goal achievement, length, and perplexity. Their method was better than their base model alone in the aspects being assessed. However, this approach requires training the model for every new goal, which can be inconvenient for the users. Another drawback to this model is it uses the sequence to sequence model in \citep{martin2017event}, which generates stories as sequences of objects encapsulating the sentence components (verb and subject) that require translation to full sentences.

\textbf{Model Fusion}. \cite{fan2018hierarchical} attempts to solving the plot controllability problem by dividing the generation process into two levels of hierarchy a premise and a story. The premise provides an overall sketch of the story, which was utilized to write the story. This fusion model combines a convolutional sequence to sequence model with a self-attention mechanism to improve generated story quality. A convolutional network first generates a writing prompt which then, becomes the input to the sequence to sequence model and guide it in generating a story conditioned on the prompt. Their model was superior in both human evaluations and perplexity scores than a traditional sequence to sequence method. Conditioning on the generated premise makes the generated story plot consistent and has an improved long-term dependency. Overall, this approach improves the shortcomings of the previous work by writing the stories directly and being conditioned for different prompts without retraining. Yet this model also has its limitations. First, it relies heavily on random sampling for the generation, which is prone to errors. Second, it suffers from text repetition in the generated stories. Lastly, the generated prompts are generic and less interesting than human written writing prompts, which often generates boring stories. 

\textbf{Plan and Write}. \cite{yao_plan-and-write_2019} proposed the Plan-and-write story generation framework. The authors leveraged some of the characteristics of symbolic planning and integrated it into a neural system. Their work improves the previous literature in that it uses the titles to generate controlled storylines rather than the auto-generated writing prompts directly. They utilize storyline planning to improve the generated stories' quality and coherence and thus control the generation. They explore several story planning strategies to see their effect on story generation. This framework takes as an input the title of the story and then generates a storyline. The storyline and the title are then used as input to control the story generation in a sequence to sequence model. They also proposed two metrics to evaluate their model, inter-story repetition, and intra-story repetition. The evaluations showed that the model is more superior to the used conditional language model baselines. Those evaluations also showed that the model suffers from several major problems: repetition, going off-topic, and logical inconsistencies. It also utilizes a sequential language model to approximate the story plot, which simplifies the structure and depth of a good story plot, suggesting that generating coherent and logical story plots is still far from being solved.  

\textbf{Generation by Interpolation}. \cite{wang2020narrative} introduced a generation-by-interpolation story generation model. While previously introduced methods require minimal human input, they still suffer from logical inconsistencies and off-topic wandering. The generation by interpolation model is designed to overcome these challenges. It is an ending-guided model that is better than storyline-guided models because, in the storyline-guided, the model can easily be misled by a very general prompt.
In contrast, an ending-guided model can use a single ending sentence to develop a good story plot. Their ending-guided method centers on conditioning the generation on the first and last sentences of the story. Where a GPT-2 model \cite{radford2019language} generates several candidates for a storyline, and then these candidates are ranked based on their coherence scores using a RoBERTa model\citep{roberta}. Then the sentence with the highest coherence with the first and last sentence is chosen and then generated. Their evaluations demonstrate the informativeness of the ending guide and the effectiveness of the coherence ranking approach. The generated stories were of higher quality and better coherence than previous state-of-the-art models. The model's human evaluations suggested that good stories' assessment needs better and deeper evaluation metrics to match how humans define an excellent story, for example, measuring how the organization of events and characters can constitute better narratives. Lastly, using a transformer-language-model-based system improved the model's coherence and repetition. However, it showed that it could not manage commonsense inference beyond a small extend and thus established the need to integrate more human knowledge into the model.

\textbf{Plot Machines}. \cite{rashkin2020plotmachines} proposed a transformer-language-model-based system that generates multi-paragraph stories conditioned on specified outlines for these stories. This model shows improvements in the narrative over the previous work. The approach utilizes memory state tracking and discourse structures to better control the generated story plot and keep track of the generated lines to maintain the coherence. The outlines are represented with an unordered list of high-level, multi-word descriptions of events occurring in the story. At every step, the model generates based on the representation of the given outline, the high-level discourse representation, the preceding story context, and the previous memory. Discourse representation is an encoding of the type of paragraph the current paragraph is, including introduction (\texttt{\_i\_}), body (\texttt{\_b\_}), and conclusion (\texttt{\_c\_}), which is appended to the outline representations at every time step. The preceding story context is the same as the hidden state vectors output by the transformer's attention blocks upon feeding generated sentences into a static GPT-2 model. Finally, the memory is a concatenated vector containing both the generated tokens and an encoded state of the story. When evaluated based on human preferences, the proposed system outperforms baseline models, including Fusion \citep{radford2018improving}, GPT-2 \citep{radford2019language}, and Grover \citep{grover} in metrics measuring logical ordering, narrative flow, and the level of repetitiveness. In PlotMachines, the conditioning of generation depended on a general outline that includes events and phrases for ease of extraction. Even with the better performance in PlotMachines, the stories can benefit from incorporating a comprehensive plot outline such as the output of an event-based planning system that can improve the generated stories' depth and interestingness. 

Narrative controllability is still an open challenge for automatic story generation. Albeit being an active research area in natural language generation, we can attribute some of its problems to the new technologies that were essentially used to improve it, which manifested after introducing neural-based systems to story generation models. As summarized in table \ref{tab:control} in appendix \ref{appendix:a}, narrative controllability approaches are typically ending-focused or storyline-focused. In the ending focused, the goal is to generate a story with a specific desired ending. An example of these such systems are \citep{tambwekar2018controllable, wang2020narrative}. Whereas in the storyline focused, the generated stories would follow an outline of the plot. \citep{rashkin2020plotmachines, yao_plan-and-write_2019, fan2018hierarchical} are examples of such systems. Both approaches reflect different controllability goals which needs to be addressed when comparing generation systems. We also notice a shift from Seq2Seq models \citep{tambwekar2018controllable,  fan2018hierarchical, yao_plan-and-write_2019} to transformer based architecture in newer models \citep{rashkin2020plotmachines, wang2020narrative}. 

After examining those solutions we notice that there are three main challenges that needs to be solved. First, controlled models generally suffer from low creativity and interestingness, which is obvious, especially in more rigid controls such as story outlines. Second, the evaluation metrics for the controllability of automatic story generation systems are neither sufficient nor unified, making it harder to evaluate and compare systems. Third, despite the controls added to the generation process, we still need to improve the coherence and logical plot generation. Those challenges are an open invitation for more research in controllability.

\subsection{Commonsense Knowledge in Story Generation}
\label{sec:cs}
Commonsense is regarded obvious to most humans\citep{cambria2011isanette}, and comprises shared
knowledge about how the world works \citep{nunberg1987position}. 
Commonsense serves as a deep understanding of language. 
Two major bottlenecks here are how to acquire commonsense knowledge and incorporate it into state-of-the-art story-telling generation systems.

\subsubsection{Benchmarks}
Before integrating commonsense knowledge into neural language models, the models often are trained on commonsense knowledge bases, datasets containing information detailing well-known facts or causal relationships. We will first introduce these benchmarks, which target commonsense.

\textbf{ConceptNet}.  ConceptNet by \citet{speer2017conceptnet} is a large semantic knowledge graph that connects words and phrases of natural language with labeled edges, describing general human knowledge and how it is expressed in natural language. The data is in form of triples of their start node, relation label, and end node. For example, the assertion that “a dog has a tail” can be represented as (dog, HasA, tail). It lays the foundation of incorporating real-world knowledge into a variety of AI projects and applications. What's more, many new benchmarks extract from ConceptNet and serve other utilities. 

%
%
\textbf{CommonsenseQA}. CommonsenseQA by \citep{talmor2019commonsenseqa} is a benchmark extracting from ConceptNet's multiple target
concepts, which have the same semantic relation, to a single source concept. 
It provides a challenging new dataset for commonsense question answering. 
Each question requires one to disambiguate a target concept from three connected concepts in ConceptNet. 
The best pre-trained LM tuned on question answering, can only get 55.9\% accuracy on CommonsenseQA, possessing important challenge for incorporating commonsense into large language model.

%
\textbf{ATOMIC}. \cite{sap2019atomic} presented ATlas Of MachIne Commonsense (ATOMIC), an atlas for commonsense knowledge with 877K textual descriptions of nine different types \textit{If-then} relations. 
Instead of capturing general commonsense knowledge like ConceptNet, ATOMIC focuses on sequences of events and the social commonsense relating to them.
The purpose of the dataset is to allow neural networks abstract commonsense inferences and make predictions on previously unseen events. 
The dataset is in the form of \texttt{<event, relation, event>} and is organized into nine categories such as xIntent (PersonX’s intention) and xEffect (effect on PersonX). For instance, “PersonX makes PersonY a birthday cake xEffect PersonX gets thanked”. 


\textbf{GLUCOSE}. ATOMIC is person centric, hence it can not be used in sentences describing events. \citet{mostafazadeh2020glucose} constructs GLUCOSE (GeneraLized and COntextualized Story Explanations), a large-scale dataset of implicit commonsense causal knowledge, which sentences can describe any event/state. Each GLUCOSE entry is organized into a story-specific causal statement paired with an inference rule generalized from the statement. Given a short story and a sentence X in the story, GLUCOSE captures ten dimensions of causal explanations related to X. GLUCOSE shares the same purpose with ATOMIC.  

\textbf{SocialIQA}. SocialIQA\citep{sap2019social} is the a large-scale benchmark for commonsense reasoning
about social situations, which provides 38k multiple choice questions. 
Each question consists of a brief context, a question about the context, and three answer options.
It covers various types of inference about people’s actions being described in situational contexts. The purpose of SocialIQA is to reason about social situations. 

There are also many other benchmarks involved in commonsense domain. \textbf{MCScript}\citep{ostermann2018mcscript} provides narrative texts and questions, collected based on script scenarios.\textbf{OpenBookQA}\citep{mihaylov2018can} is a question answering dataset, modeled after open
book exams for assessing human understanding of a subject.
\textbf{Cosmos QA}\citep{huang2019cosmos} provides 35k problems with multiple-choice, which
require commonsense-based reading comprehension.

What's more, technique of generating commonsense datasets are also developed. For example, \citet{davison2019commonsense} proposed a method for generating commonsense knowledge by transforming relational triples into masked sentences, and then using a large, pre-trained bidirectional language model to rank a triple’s validity by the estimated pointwise mutual information between the two entities. 
\citet{schwartz2017effect} and \citet{trinh2018simple} demonstrate a
similar approach to using language models for tasks requiring commonsense, such as the Story Cloze Task and the Winograd Schema Challenge, respectively \citep{mostafazadeh2016corpus, levesque2012winograd}.

\subsubsection{Frameworks}

Three ways of applying these benchmarks on commonsense story generation are (1) fine-tuning pre-trained language models (LM) on commonsense benchmarks, (2) perceptions of causality after generating stories, and (3) incorporating benchmarks into language models encoding.

An intuition is to utilize commonsense knowledge is to train language model on commonsense datasets. 
\citet{yang2019enhancing} integrates external commonsense knowledge to BERT \citep{devlin2019bert} to enhance language representation for reading comprehension. 
\citet{guan2020knowledge} fine-tuned GPT-2\citep{radford2019language} on on knowledge-augmented data, ATOMIC and ConceptNet, for a better performance for commonsense story generation. They firstly transform ConceptNet and ATOMIC into readable natural language sentences and then post-trained on these transformed sentences by minimizing the negative likelihood of predicting the next token. \citet{mao2019improving} and \cite{guan2020knowledge} also fine-tuned GPT-2 on ConceptNet and the
BookCorpus\citep{kiros2015skip}. 
They achieve a less perplexity and higher BLEU score,
however, these knowledge-enhanced pre-training model for commonsense story generation are still far from generating stories with long-range coherence.

Instead of directly training language models on commonsense datasets, which improves LM's logicality and grammaticality,
an alternative of incorporating commonsense into language model is to analyze perceptions of causality or overall story quality.

\cite{bosselut2019comet} extended upon the work ATOMIC by \citet{sap2019atomic} and ConceptNet by \citet{speer2017conceptnet} and trained a GPT model \citep{radford2018improving} on commonsense knowledge tuples, in the format of \texttt{<phrase subject, relationship, phrase object>}. The resulting model, \textbf{COMeT}, is capable of generating new commonsense triples on novel phrases. 
With this feature, automatic generated story can be evaluated easily. The model has been proven to be efficient in learning commonsense knowledge tuples, as in humans deem most COMeT-generated triples from novel phrases to be correct. It provides a easy way of making inferece on generated text.
However, it is Sentence-level Commonsense inferences, which is only able to deal with short sentences, within 18 tokens. Story generation is usually in need of a paragraph-level commonsense inference because combining with context, the inference could be completely different.

In order to incorporates paragraph-level information to generate coherent commonsense inferences from narratives, \citet{gabriel2020paragraph} proposed a discourse-aware model \textbf{PARA-COMeT}. PARA-COMeT firstly created commonsense datasets by (1) using COMeT to provides inference on sentences in ROCStories corpus \citep{mostafazadeh2016corpus} and (2) transform inference into natural language by human-written templates, (3) then filter out those with low coherence with narrative. PARA-COMeT consists of (1) a memory-less model, focusing on extracting semantic knowledge from the context, and (2) a model augmented with recurrent memory, used for incorporating episodic knowledge. Compared with COMeT, PARA-COMeT demonstrated the effectiveness of generating more implicit and novel discourse-aware inferences in paragraph level. 

\citet{ammanabrolu2020automated} also developed proposed \textbf{C}ausal, \textbf{C}ommonsense \textbf{P}ot \textbf{O}rdering(CCPO) on COMeT. CCPO establishs plot points by (1) extracting all the coreference clusters from a given textual story plot using a pre-trained neural coreference resolution model\citep{clark2016deep}, and (2)extract a set of <subject, relation, object> triples from the story text using OpenIE\citep{angeli2015leveraging}. Then a plot graph between each two plot points is generated by keep recursively querying commonsense inference on these two plot points. The automatic story is generated based on the plot graphs. CCPO successfully improves perceptions of local and global coherence in terms of causality, however its performance is restricted by commonsense inference models.

Another common method is incorporating commonsense knowledge graph into the model encoding process.
\citet{guan2019story} incorporates commonsense knowledge graph by applying features from ConceptNet\citep{speer2017conceptnet} and graph attention\citep{velivckovic2018graph} on building knowledge context vectors to represent the graph. 
They significantly improve the ability of neural networks to predict the end of a story.
\citet{mihaylov2018knowledgeable} also incorporate external
commonsense knowledge into a neural cloze-style reading comprehension model.

\subsection{Other Challenges in Story Generation}
\label{sec:other}
There are issues in the story generation field that are yet to be heavily researched upon. The current emphasis of mainstream story generation research is to produce narratives with reasonable structures and plots and less on the cherries on top: fascinating and driven characters, consistent styles, and creative language and plot. Some researchers have ventured potential approaches to these currently outstanding problems, as detailed below.

\subsubsection{Characters and Entities}
How characters are motivated and interact with each other influence the progression of a story. Different approaches have been taken to model how focusing on characters can produce higher-quality generated narratives, some from the perspective of character affect, and some from entity representation in narrative generation.

\textbf{\textsc{EnGen}} \cite{clark-etal-2018-neural} presented an entity-based generation model \textsc{EnGen}, which produces narratives relying on: (1) the current sentence; (2) the previous sentence, encoded by a Seq2Seq model (S2SA); (3) dynamic, up-to-date representations of all the entities in the narrative. The entity representation vectors are based on EntityNLM \citep{ji2017dynamic}, and the vectors are updated every time their corresponding entities are mentioned. The model was evaluated on a series of tasks, including a novel mention generation task, where the model fills a slot with all previous mentions of entities, including coreferences. Similarly, the automated sentence selection task examines \textsc{EnGen}'s ability to distinguish between the ground truth continuation sentence and a distraction sentence. \textsc{EnGen} is able to out-perform both S2SA and EntityNLM for these tasks. Another task involved Mechanical Turk workers reading sentences generated by both \textsc{EnGen} and S2SA on the same prompts and deciding which continuation is more fluent. Out of the 50 prompt passages, Turkers preferred the \textsc{EnGen} stories for 27 of them, and S2SA for the rest 23, although most of the human evaluations yield similar results between the two models. Incorporating character or entity information into the context for generation can improve model performance on some automated and human-evaluated tasks. The authors contended that this design improves the fluency of the generated texts. However, the lengths of the generated segments for the human-evaluation task are very short, usually fragments of sentences. Therefore, it is unlikely that these generated texts help propel the plot. Furthermore, the paper does not indicate how the entity representations model character interactions and how these interactions contribute to the stories.

\textbf{Using Character Affinities} A dive into character interactions in particular is detailed in \cite{mendez2016use}, where the authors attempted to model character interactions using numerical affinity values. Character relationships are categorized into four types: foe (lowest affinity), indifferent (medium affinity), friend (higher affinity), and mate (highest affinity). The system consists of a Director Agent, which sets up the environment for interactions to occur, and a set of Character Agents, each representing a character. The authors defines that each Character Agent interacts with the character's foes, friends, and mates. Actions pertinent to different interactions are templated using defined interaction protocols and are relatively restricted in terms of scope. These actions are independent and can be added upon each other to alter the affinity values. The primary parameter of concern in this model is the affinity between characters, a factor related to character emotions. Although this modeling approach has been suggested for narrative generation, the authors did not provide examples of stories generated using this character affinity model. Instead, the authors presented affinity changes for different Character Agents in the story to illustrate how different affinity threshold values for foe interactions affect the affinity evolution in the narratives. The model might be considered useful for modeling character interactions, yet the effect affinity changes have on the story plot remains unclear.

\textbf{EC-CLF} \cite{brahman2020modeling} proposed a method for story generation conditioned on emotion arc of the protagonist by using reinforcement learning to train a GPT-2 model. The authors suggested two emotion consistency rewards: EC-EM and EC-CLF. EC-EM calculates how well the generated story aligns with the given arc using character reaction inferences from COMET \citep{bosselut2019comet}; it is a modified Levensthtein distance that considers the cosine similarities between words from the given arc and the COMET inferences. EC-CLF, on the other hand, involves training a BERT \citep{devlin2019bert} classifier to identify the emotion in the generated sentences; the reward value is the probability of the desired emotions throughout the narrative from the classifier head. For human-evaluated tasks such as assessing emotion faithfulness and content quality, RL-CLF (GPT-2 trained with EC-CLF reward) outperformed baselines including GPT-2 trained with the emotion arc as an additional input to the narrative examples (EmoSup) and GPT-2 trained on the reward function EC-EM. This work augmented current state-of-the-art models with the ability to generate narratives with the protagonist's emotion changes following a specified emotion arc. It is an example of how character emotions can be used to inform story progression and improve narrative quality. Despite the enhancement of generation quality, the model still only focuses on one character instead of interactions between characters.

\textbf{SRL + Entity} \cite{fan2019strategies} generated action-driven narratives by adapting the following pipeline: (1) based on the prompt given, produce an action plan with where all entities are represented with placeholder tags; (2) create an entity-anonymized story from the action plan; (3) output the full story after replacing the anonymized, generalized entities with natural language entities. Every entry in the action sequence consists of a predicate, which is a verb, and a series of arguments, which are the entities involved in the action. This representation allows models to learn more in-depth and generalizable relationships between different verbs and characters. A convolutional Seq2Seq model is trained on the prompts from the \textsc{WritingPrompts} dataset \citep{fan2018hierarchical} and their corresponding action sequences. The network has an attention head dedicated to past verbs to improve verb diversity in generations.Human preference studies showed that the novel model generated more coherent narratives than the \textit{Fusion} model from \cite{fan2018hierarchical}; additionally, the new model had more diversity in the generated verbs. The technique of abstraction and generalization can be proven useful in the story generation process, since abstractions reveal more widely-applicable rules in storytelling. Again, it is not clear if character interactions are implicitly learned by the models in this work, therefore further investigation would be required to determine if this work is suitable for multi-agent narrative generation.

In this section, we examine four works in the sub-field of character and entity-focused automated narrative generation. Generally, representing entities in certain format can improve the quality of the plotline, and character emotions can help inform the story generation process. Interactions between multiple characters are currently not the focus of the field, but it should be for potential future research.

\subsubsection{Creativity}
Creativity in human-authored narratives manifests in ways including figures of speech, character traits, and the environment for the narrative to occur in. \cite{lara-improv} developed a system for improvisational interactive storytelling based on a plot graph as a general guideline for the generated storyline. Recent introduction to transformer-based language models has inspired people generating novel contents using these language models \footnote{\url{https://www.gwern.net/GPT-3}}, including using GPT-2 to generate fantasy descriptions with explicit subjects and weblinks \citep{austin2019book}. Nonetheless, there has still not been much specific research into further improving the creativity of transformer-based language models.

\section{Conclusion and Future Work}
This survey discussed several directions in automatic story generation research and their respective challenges, and summarized research attempts at solving them. The research in automatic story generation is far from done. As with every new technology, new challenges arise. With automated story generation, such challenges include controlling the story content, commonsense knowledge, inferring reasonable character actions, and creativity.  This survey provides a dive into some of these active research problems. This survey's value is that it is a good starting point for researchers who want to learn more about the domain and the current state-of-the-art solutions for several story generation challenges. 

In Section~\ref{sec:co}, we summarized a few approaches addressing the problem of story generation controllability. We noticed that the papers we reviewed shared one of two goals, either controlling the story outline or controlling the story ending. We also observed an emerging trend towards using transformer-based language models for story generation. 

In Section~\ref{sec:cs}, we introduced methods to incorporate commonsense knowledge into story generation systems and frameworks with such integrated commonsense knowledge. Frequent approaches include: (1) Fine-tuning on commonsense datasets, (2) analyzing perceptions of causality and (3) incorporating commonsense knowledge graph into encoders. These methods are able to increase the overall story quality. However, no methods can ensure the generation of reasonable and coherent stories. One potential path to major improvements in this area would be to combine all of these different approaches.

In Section ~\ref{sec:other}, we provided insight into some less-researched areas at the moment, including characters in generated narratives and the creativity of generated stories. Incorporating representations of entities into the generation process seems to improve the coherence of the plot, and character affect can help navigate the generation space as well. Extending the work in character affect from single character to multi characters can perhaps further enhance the generated narratives. There has not been much emphasis on the creativity of generated texts.

Additionally, we highlight a few future research problems that are worth exploring:

\begin{enumerate}
\item In the controllability systems we examined, we noticed that the stories become less interesting when the generation process is more controlled. There is a trade-off between narrative creativity and structural coherence of narratives. 

\item The evaluation metrics used are generally the metrics used for other natural language generation tasks such as BLEU, perplexity, and ROUGE. Those metrics are weak and do not perform well for this task. Moreover, the story generation domain needs different metrics to capture story-specific characteristics. Such as measures for creativity and interestingness. Besides, there is a need to develop more robust and unified metrics to facilitate comparisons between systems.

\item The problems of plot incoherence and illogical plot generation are far from being solved. Both are still very active research areas and can be an interesting future research direction.

\item Instead of sentence-level and paragraph-level commonsense inference, a story-level commonsense inference could increase the accuracy of inference and provides a better tool for generating a more logic coherent story.
\end{enumerate}


\newpage
\bibliographystyle{acl_natbib}
\bibliography{references}

\begin{thebibliography}{57}
\expandafter\ifx\csname natexlab\endcsname\relax\def\natexlab#1{#1}\fi

\bibitem[{Ammanabrolu et~al.(2020)Ammanabrolu, Cheung, Broniec, and
  Riedl}]{ammanabrolu2020automated}
Prithviraj Ammanabrolu, Wesley Cheung, William Broniec, and Mark~O Riedl. 2020.
\newblock Automated storytelling via causal, commonsense plot ordering.
\newblock \emph{arXiv preprint arXiv:2009.00829}.

\bibitem[{Angeli et~al.(2015)Angeli, Premkumar, and
  Manning}]{angeli2015leveraging}
Gabor Angeli, Melvin Jose~Johnson Premkumar, and Christopher~D Manning. 2015.
\newblock Leveraging linguistic structure for open domain information
  extraction.
\newblock In \emph{Proceedings of the 53rd Annual Meeting of the Association
  for Computational Linguistics and the 7th International Joint Conference on
  Natural Language Processing (Volume 1: Long Papers)}, pages 344--354.

\bibitem[{Austin(2019)}]{austin2019book}
John Austin. 2019.
\newblock The book of endless history: Authorial use of gpt2 for interactive
  storytelling.
\newblock In \emph{International Conference on Interactive Digital
  Storytelling}, pages 429--432. Springer.

\bibitem[{Bosselut et~al.(2019)Bosselut, Rashkin, Sap, Malaviya, Celikyilmaz,
  and Choi}]{bosselut2019comet}
Antoine Bosselut, Hannah Rashkin, Maarten Sap, Chaitanya Malaviya, Asli
  Celikyilmaz, and Yejin Choi. 2019.
\newblock Comet: Commonsense transformers for automatic knowledge graph
  construction.
\newblock pages 4762--4779.

\bibitem[{Brahman and Chaturvedi(2020)}]{brahman2020modeling}
Faeze Brahman and Snigdha Chaturvedi. 2020.
\newblock Modeling protagonist emotions for emotion-aware storytelling.
\newblock \emph{arXiv preprint arXiv:2010.06822}.

\bibitem[{Cambria et~al.(2011)Cambria, Song, Wang, and
  Hussain}]{cambria2011isanette}
Erik Cambria, Yangqiu Song, Haixun Wang, and Amir Hussain. 2011.
\newblock Isanette: A common and common sense knowledge base for opinion
  mining.
\newblock In \emph{2011 IEEE 11th International Conference on Data Mining
  Workshops}, pages 315--322. IEEE.

\bibitem[{Cavazza and Pizzi(2006)}]{cavazza_narratology_2006}
Marc Cavazza and David Pizzi. 2006.
\newblock \href {https://doi.org/10.1007/11944577_7} {Narratology for
  {Interactive} {Storytelling}: {A} {Critical} {Introduction}}.
\newblock In \emph{Technologies for {Interactive} {Digital} {Storytelling} and
  {Entertainment}}, Lecture {Notes} in {Computer} {Science}, pages 72--83,
  Berlin, Heidelberg. Springer.

\bibitem[{Clark et~al.(2018)Clark, Ji, and Smith}]{clark-etal-2018-neural}
Elizabeth Clark, Yangfeng Ji, and Noah~A. Smith. 2018.
\newblock Neural text generation in stories using entity representations as
  context.
\newblock In \emph{Proceedings of the 2018 Conference of the North {A}merican
  Chapter of the Association for Computational Linguistics: Human Language
  Technologies, Volume 1 (Long Papers)}.

\bibitem[{Clark and Manning(2016)}]{clark2016deep}
Kevin Clark and Christopher~D Manning. 2016.
\newblock Deep reinforcement learning for mention-ranking coreference models.
\newblock In \emph{Proceedings of the 2016 Conference on Empirical Methods in
  Natural Language Processing}, pages 2256--2262.

\bibitem[{Davison et~al.(2019)Davison, Feldman, and
  Rush}]{davison2019commonsense}
Joe Davison, Joshua Feldman, and Alexander~M Rush. 2019.
\newblock Commonsense knowledge mining from pretrained models.
\newblock In \emph{Proceedings of the 2019 Conference on Empirical Methods in
  Natural Language Processing and the 9th International Joint Conference on
  Natural Language Processing (EMNLP-IJCNLP)}, pages 1173--1178.

\bibitem[{Devlin et~al.(2019)Devlin, Chang, Lee, and
  Toutanova}]{devlin2019bert}
Jacob Devlin, Ming-Wei Chang, Kenton Lee, and Kristina Toutanova. 2019.
\newblock Bert: Pre-training of deep bidirectional transformers for language
  understanding.
\newblock In \emph{Proceedings of the 2019 Conference of the North American
  Chapter of the Association for Computational Linguistics: Human Language
  Technologies, Volume 1 (Long and Short Papers)}, pages 4171--4186.

\bibitem[{Fan et~al.(2018)Fan, Lewis, and Dauphin}]{fan2018hierarchical}
Angela Fan, Mike Lewis, and Yann Dauphin. 2018.
\newblock Hierarchical neural story generation.
\newblock pages 889--898.

\bibitem[{Fan et~al.(2019)Fan, Lewis, and Dauphin}]{fan2019strategies}
Angela Fan, Mike Lewis, and Yann Dauphin. 2019.
\newblock Strategies for structuring story generation.
\newblock In \emph{Proceedings of the 57th Annual Meeting of the Association
  for Computational Linguistics}, pages 2650--2660.

\bibitem[{Gabriel et~al.(2020)Gabriel, Bhagavatula, Shwartz, Bras, Forbes, and
  Choi}]{gabriel2020paragraph}
Saadia Gabriel, Chandra Bhagavatula, Vered Shwartz, Ronan~Le Bras, Maxwell
  Forbes, and Yejin Choi. 2020.
\newblock Paragraph-level commonsense transformers with recurrent memory.
\newblock \emph{arXiv preprint arXiv:2010.01486}.

\bibitem[{Gerv{\'a}s(2009)}]{gervas2009computational}
Pablo Gerv{\'a}s. 2009.
\newblock Computational approaches to storytelling and creativity.
\newblock \emph{AI Magazine}, 30(3):49--49.

\bibitem[{Guan et~al.(2020)Guan, Huang, Zhao, Zhu, and
  Huang}]{guan2020knowledge}
Jian Guan, Fei Huang, Zhihao Zhao, Xiaoyan Zhu, and Minlie Huang. 2020.
\newblock A knowledge-enhanced pretraining model for commonsense story
  generation.
\newblock \emph{Transactions of the Association for Computational Linguistics},
  8:93--108.

\bibitem[{Guan et~al.(2019)Guan, Wang, and Huang}]{guan2019story}
Jian Guan, Yansen Wang, and Minlie Huang. 2019.
\newblock Story ending generation with incremental encoding and commonsense
  knowledge.
\newblock In \emph{Proceedings of the AAAI Conference on Artificial
  Intelligence}, volume~33, pages 6473--6480.

\bibitem[{Hou et~al.(2019)Hou, Zhou, Zhou, Sun, and Xuanyuan}]{survey_deep}
Chenglong Hou, Chensong Zhou, Kun Zhou, Jinan Sun, and Sisi Xuanyuan. 2019.
\newblock A survey of deep learning applied to story generation.
\newblock In \emph{Smart Computing and Communication}, pages 1--10, Cham.
  Springer International Publishing.

\bibitem[{Huang et~al.(2019)Huang, Le~Bras, Bhagavatula, and
  Choi}]{huang2019cosmos}
Lifu Huang, Ronan Le~Bras, Chandra Bhagavatula, and Yejin Choi. 2019.
\newblock Cosmos qa: Machine reading comprehension with contextual commonsense
  reasoning.
\newblock In \emph{Proceedings of the 2019 Conference on Empirical Methods in
  Natural Language Processing and the 9th International Joint Conference on
  Natural Language Processing (EMNLP-IJCNLP)}, pages 2391--2401.

\bibitem[{Ji et~al.(2017)Ji, Tan, Martschat, Choi, and Smith}]{ji2017dynamic}
Yangfeng Ji, Chenhao Tan, Sebastian Martschat, Yejin Choi, and Noah~A Smith.
  2017.
\newblock Dynamic entity representations in neural language models.
\newblock In \emph{Proceedings of the 2017 Conference on Empirical Methods in
  Natural Language Processing}, pages 1830--1839.

\bibitem[{Kiros et~al.(2015)Kiros, Zhu, Salakhutdinov, Zemel, Urtasun,
  Torralba, and Fidler}]{kiros2015skip}
Ryan Kiros, Yukun Zhu, Russ~R Salakhutdinov, Richard Zemel, Raquel Urtasun,
  Antonio Torralba, and Sanja Fidler. 2015.
\newblock Skip-thought vectors.
\newblock In \emph{Advances in neural information processing systems}, pages
  3294--3302.

\bibitem[{{Kybartas} and {Bidarra}(2017)}]{7439785}
B.~{Kybartas} and R.~{Bidarra}. 2017.
\newblock A survey on story generation techniques for authoring computational
  narratives.
\newblock \emph{IEEE Transactions on Computational Intelligence and AI in
  Games}, 9(3):239--253.

\bibitem[{Lehnert et~al.(1983)Lehnert, Dyer, Johnson, Yang, and
  Harley}]{lehnert1983boris}
Wendy~G Lehnert, Michael~G Dyer, Peter~N Johnson, CJ~Yang, and Steve Harley.
  1983.
\newblock Boris—an experiment in in-depth understanding of narratives.
\newblock \emph{Artificial intelligence}, 20(1):15--62.

\bibitem[{Levesque et~al.(2012)Levesque, Davis, and
  Morgenstern}]{levesque2012winograd}
Hector Levesque, Ernest Davis, and Leora Morgenstern. 2012.
\newblock The winograd schema challenge.
\newblock In \emph{Thirteenth International Conference on the Principles of
  Knowledge Representation and Reasoning}. Citeseer.

\bibitem[{Liu et~al.(2019)Liu, Ott, Goyal, Du, Joshi, Chen, Levy, Lewis,
  Zettlemoyer, and Stoyanov}]{roberta}
Yinhan Liu, Myle Ott, Naman Goyal, Jingfei Du, Mandar Joshi, Danqi Chen, Omer
  Levy, Mike Lewis, Luke Zettlemoyer, and Veselin Stoyanov. 2019.
\newblock Roberta: A robustly optimized bert pretraining approach.
\newblock \emph{arXiv preprint arXiv:1907.11692}.

\bibitem[{Luo et~al.(2015)Luo, Cai, Zhou, Lees, and Yin}]{luo_review_2015}
Linbo Luo, Wentong Cai, Suiping Zhou, Michael Lees, and Haiyan Yin. 2015.
\newblock \href {https://doi.org/10.1177/0037549714566722} {A review of
  interactive narrative systems and technologies: a training perspective}.
\newblock \emph{SIMULATION}, 91(2):126--147.
\newblock Publisher: SAGE Publications Ltd STM.

\bibitem[{Mao et~al.(2019)Mao, Majumder, McAuley, and
  Cottrell}]{mao2019improving}
Huanru~Henry Mao, Bodhisattwa~Prasad Majumder, Julian McAuley, and Garrison
  Cottrell. 2019.
\newblock Improving neural story generation by targeted common sense grounding.
\newblock In \emph{Proceedings of the 2019 Conference on Empirical Methods in
  Natural Language Processing and the 9th International Joint Conference on
  Natural Language Processing (EMNLP-IJCNLP)}, pages 5990--5995.

\bibitem[{Martin et~al.(2017)Martin, Ammanabrolu, Wang, Hancock, Singh,
  Harrison, and Riedl}]{martin2017event}
Lara~J Martin, Prithviraj Ammanabrolu, Xinyu Wang, William Hancock, Shruti
  Singh, Brent Harrison, and Mark~O Riedl. 2017.
\newblock Event representations for automated story generation with deep neural
  nets.
\newblock \emph{arXiv preprint arXiv:1706.01331}.

\bibitem[{Martin et~al.(2016)Martin, Harrison, and Riedl}]{lara-improv}
Lara~J. Martin, Brent Harrison, and Mark~O. Riedl. 2016.
\newblock Improvisational computational storytelling in open worlds.
\newblock In \emph{Interactive Storytelling}, pages 73--84, Cham. Springer
  International Publishing.

\bibitem[{M{\'e}ndez et~al.(2016)M{\'e}ndez, Gerv{\'a}s, and
  Le{\'o}n}]{mendez2016use}
Gonzalo M{\'e}ndez, Pablo Gerv{\'a}s, and Carlos Le{\'o}n. 2016.
\newblock On the use of character affinities for story plot generation.
\newblock In \emph{Knowledge, Information and Creativity Support Systems},
  pages 211--225. Springer.

\bibitem[{Mihaylov et~al.(2018)Mihaylov, Clark, Khot, and
  Sabharwal}]{mihaylov2018can}
Todor Mihaylov, Peter Clark, Tushar Khot, and Ashish Sabharwal. 2018.
\newblock Can a suit of armor conduct electricity? a new dataset for open book
  question answering.
\newblock In \emph{Proceedings of the 2018 Conference on Empirical Methods in
  Natural Language Processing}, pages 2381--2391.

\bibitem[{Mihaylov and Frank(2018)}]{mihaylov2018knowledgeable}
Todor Mihaylov and Anette Frank. 2018.
\newblock Knowledgeable reader: Enhancing cloze-style reading comprehension
  with external commonsense knowledge.
\newblock In \emph{Proceedings of the 56th Annual Meeting of the Association
  for Computational Linguistics (Volume 1: Long Papers)}, pages 821--832.

\bibitem[{Mostafazadeh et~al.(2016)Mostafazadeh, Chambers, He, Parikh, Batra,
  Vanderwende, Kohli, and Allen}]{mostafazadeh2016corpus}
Nasrin Mostafazadeh, Nathanael Chambers, Xiaodong He, Devi Parikh, Dhruv Batra,
  Lucy Vanderwende, Pushmeet Kohli, and James Allen. 2016.
\newblock A corpus and evaluation framework for deeper understanding of
  commonsense stories.
\newblock \emph{arXiv preprint arXiv:1604.01696}.

\bibitem[{Mostafazadeh et~al.(2020)Mostafazadeh, Kalyanpur, Moon, Buchanan,
  Berkowitz, Biran, and Chu-Carroll}]{mostafazadeh2020glucose}
Nasrin Mostafazadeh, Aditya Kalyanpur, Lori Moon, David Buchanan, Lauren
  Berkowitz, Or~Biran, and Jennifer Chu-Carroll. 2020.
\newblock Glucose: Generalized and contextualized story explanations.
\newblock \emph{arXiv preprint arXiv:2009.07758}.

\bibitem[{Nunberg(1987)}]{nunberg1987position}
Geoffrey Nunberg. 1987.
\newblock Position paper on common-sense and formal semantics.
\newblock In \emph{Theoretical Issues in Natural Language Processing 3}.

\bibitem[{Ostermann et~al.(2018)Ostermann, Modi, Roth, Thater, and
  Pinkal}]{ostermann2018mcscript}
Simon Ostermann, Ashutosh Modi, Michael Roth, Stefan Thater, and Manfred
  Pinkal. 2018.
\newblock Mcscript: A novel dataset for assessing machine comprehension using
  script knowledge.
\newblock In \emph{Proceedings of the Eleventh International Conference on
  Language Resources and Evaluation (LREC 2018)}.

\bibitem[{Radford et~al.(2018)Radford, Narasimhan, Salimans, and
  Sutskever}]{radford2018improving}
Alec Radford, Karthik Narasimhan, Tim Salimans, and Ilya Sutskever. 2018.
\newblock Improving language understanding by generative pre-training.

\bibitem[{Radford et~al.(2019)Radford, Wu, Child, Luan, Amodei, and
  Sutskever}]{radford2019language}
Alec Radford, Jeffrey Wu, Rewon Child, David Luan, Dario Amodei, and Ilya
  Sutskever. 2019.
\newblock Language models are unsupervised multitask learners.
\newblock \emph{OpenAI Blog}, 1(8).

\bibitem[{Rashkin et~al.(2020)Rashkin, Celikyilmaz, Choi, and
  Gao}]{rashkin2020plotmachines}
Hannah Rashkin, Asli Celikyilmaz, Yejin Choi, and Jianfeng Gao. 2020.
\newblock Plotmachines: Outline-conditioned generation with dynamic plot state
  tracking.
\newblock \emph{arXiv preprint arXiv:2004.14967}.

\bibitem[{Reeves(1991)}]{reeves1991computational}
John~Fairbanks Reeves. 1991.
\newblock Computational morality: A process model of belief conflict and
  resolution for story understanding.

\bibitem[{Riedl(2016)}]{riedl_computational_2016}
Mark~O. Riedl. 2016.
\newblock \href {http://arxiv.org/abs/1602.06484} {Computational {Narrative}
  {Intelligence}: {A} {Human}-{Centered} {Goal} for {Artificial}
  {Intelligence}}.
\newblock \emph{arXiv:1602.06484 [cs]}.
\newblock ArXiv: 1602.06484.

\bibitem[{Riedl and Bulitko(2013)}]{riedl2013interactive}
Mark~Owen Riedl and Vadim Bulitko. 2013.
\newblock Interactive narrative: An intelligent systems approach.
\newblock \emph{Ai Magazine}, 34(1):67--67.

\bibitem[{Roberts and Isbell(2008)}]{roberts_survey_2008}
David~L Roberts and Charles~L Isbell. 2008.
\newblock A {Survey} and {Qualitative} {Analysis} of {Recent} {Advances} in
  {Drama} {Management}.
\newblock page~15.

\bibitem[{Sap et~al.(2019{\natexlab{a}})Sap, Le~Bras, Allaway, Bhagavatula,
  Lourie, Rashkin, Roof, Smith, and Choi}]{sap2019atomic}
Maarten Sap, Ronan Le~Bras, Emily Allaway, Chandra Bhagavatula, Nicholas
  Lourie, Hannah Rashkin, Brendan Roof, Noah~A Smith, and Yejin Choi.
  2019{\natexlab{a}}.
\newblock Atomic: An atlas of machine commonsense for if-then reasoning.
\newblock In \emph{Proceedings of the AAAI Conference on Artificial
  Intelligence}, volume~33, pages 3027--3035.

\bibitem[{Sap et~al.(2019{\natexlab{b}})Sap, Rashkin, Chen, Le~Bras, and
  Choi}]{sap2019social}
Maarten Sap, Hannah Rashkin, Derek Chen, Ronan Le~Bras, and Yejin Choi.
  2019{\natexlab{b}}.
\newblock Social iqa: Commonsense reasoning about social interactions.
\newblock In \emph{Proceedings of the 2019 Conference on Empirical Methods in
  Natural Language Processing and the 9th International Joint Conference on
  Natural Language Processing (EMNLP-IJCNLP)}, pages 4453--4463.

\bibitem[{Schwartz et~al.(2017)Schwartz, Sap, Konstas, Zilles, Choi, and
  Smith}]{schwartz2017effect}
Roy Schwartz, Maarten Sap, Ioannis Konstas, Leila Zilles, Yejin Choi, and
  Noah~A Smith. 2017.
\newblock The effect of different writing tasks on linguistic style: A case
  study of the roc story cloze task.
\newblock In \emph{Proceedings of the 21st Conference on Computational Natural
  Language Learning (CoNLL 2017)}, pages 15--25.

\bibitem[{Speer et~al.(2017)Speer, Chin, and Havasi}]{speer2017conceptnet}
Robyn Speer, Joshua Chin, and Catherine Havasi. 2017.
\newblock Conceptnet 5.5: an open multilingual graph of general knowledge.
\newblock In \emph{Proceedings of the Thirty-First AAAI Conference on
  Artificial Intelligence}, pages 4444--4451.

\bibitem[{Suzuki et~al.(2018)Suzuki, Feliú-Mójer, Hasson, Yehuda, and
  Zarate}]{suzuki_dialogues_2018}
Wendy~A. Suzuki, Mónica~I. Feliú-Mójer, Uri Hasson, Rachel Yehuda, and
  Jean~Mary Zarate. 2018.
\newblock \href {https://doi.org/10.1523/JNEUROSCI.1942-18.2018} {Dialogues:
  {The} {Science} and {Power} of {Storytelling}}.
\newblock \emph{The Journal of Neuroscience}, 38(44):9468--9470.

\bibitem[{Talmor et~al.(2019)Talmor, Herzig, Lourie, and
  Berant}]{talmor2019commonsenseqa}
Alon Talmor, Jonathan Herzig, Nicholas Lourie, and Jonathan Berant. 2019.
\newblock Commonsenseqa: A question answering challenge targeting commonsense
  knowledge.
\newblock In \emph{Proceedings of the 2019 Conference of the North American
  Chapter of the Association for Computational Linguistics: Human Language
  Technologies, Volume 1 (Long and Short Papers)}, pages 4149--4158.

\bibitem[{Tambwekar et~al.(2019)Tambwekar, Dhuliawala, Martin, Mehta, Harrison,
  and Riedl}]{tambwekar2018controllable}
Pradyumna Tambwekar, Murtaza Dhuliawala, Lara~J Martin, Animesh Mehta, Brent
  Harrison, and Mark~O Riedl. 2019.
\newblock Controllable neural story plot generation via reward shaping.
\newblock pages 5982--5988.

\bibitem[{Trinh and Le(2018)}]{trinh2018simple}
Trieu~H Trinh and Quoc~V Le. 2018.
\newblock A simple method for commonsense reasoning.
\newblock \emph{arXiv preprint arXiv:1806.02847}.

\bibitem[{Veli{\v{c}}kovi{\'c} et~al.(2018)Veli{\v{c}}kovi{\'c}, Cucurull,
  Casanova, Romero, Li{\`o}, and Bengio}]{velivckovic2018graph}
Petar Veli{\v{c}}kovi{\'c}, Guillem Cucurull, Arantxa Casanova, Adriana Romero,
  Pietro Li{\`o}, and Yoshua Bengio. 2018.
\newblock Graph attention networks.
\newblock In \emph{International Conference on Learning Representations}.

\bibitem[{Wang et~al.(2020)Wang, Durrett, and Erk}]{wang2020narrative}
Su~Wang, Greg Durrett, and Katrin Erk. 2020.
\newblock Narrative interpolation for generating and understanding stories.
\newblock \emph{arXiv preprint arXiv:2008.07466}.

\bibitem[{Yang et~al.(2019)Yang, Wang, Liu, Liu, Lyu, Wu, She, and
  Li}]{yang2019enhancing}
An~Yang, Quan Wang, Jing Liu, Kai Liu, Yajuan Lyu, Hua Wu, Qiaoqiao She, and
  Sujian Li. 2019.
\newblock Enhancing pre-trained language representations with rich knowledge
  for machine reading comprehension.
\newblock In \emph{Proceedings of the 57th Annual Meeting of the Association
  for Computational Linguistics}, pages 2346--2357.

\bibitem[{Yao et~al.(2019)Yao, Peng, Weischedel, Knight, Zhao, and
  Yan}]{yao_plan-and-write_2019}
Lili Yao, Nanyun Peng, Ralph Weischedel, Kevin Knight, Dongyan Zhao, and Rui
  Yan. 2019.
\newblock \href {https://doi.org/10.1609/aaai.v33i01.33017378}
  {Plan-and-{Write}: {Towards} {Better} {Automatic} {Storytelling}}.
\newblock \emph{Proceedings of the AAAI Conference on Artificial Intelligence},
  33:7378--7385.

\bibitem[{Young et~al.(2013)Young, Ware, Cassell, and
  Robertson}]{young_plans_2013}
R~Michael Young, Stephen Ware, Brad Cassell, and Justus Robertson. 2013.
\newblock Plans and {Planning} in {Narrative} {Generation}: {A} {Review} of
  {Plan}-{Based} {Approaches} to the {Generation} of {Story}, {Dis}- course and
  {Interactivity} in {Narratives}.
\newblock page~24.

\bibitem[{Zellers et~al.(2019)Zellers, Holtzman, Rashkin, Bisk, Farhadi,
  Roesner, and Choi}]{grover}
Rowan Zellers, Ari Holtzman, Hannah Rashkin, Yonatan Bisk, Ali Farhadi,
  Franziska Roesner, and Yejin Choi. 2019.
\newblock Defending against neural fake news.
\newblock In \emph{Advances in Neural Information Processing Systems}, pages
  9054--9065.

\end{thebibliography}

\newpage
\begin{appendices}
\section{Controllability Approaches} \label{appendix:a}

\begin{table}[h!]
\caption{\label{tab:control} Summary of controllability approaches}
\resizebox{\textwidth}{!}{%
\begin{tabular}{|l|p{6cm}|l|l|}
\hline
Model/System & Architecture & Condition & Goal \\ \hline
Reinforcement Learning & Reinforcement Learning on a Seq2Seq model & Goal Event & Generate a specific ending \\ \hline
Model Fusion & Generation on two levels: CNN to generate prompt,  Seq2Seq to generate story from prompt & Generated Prompt & Generate with a storyline \\ \hline
Plan and Write & Two Seq2Seq models for plot and story generation & Title & Generate with a storyline \\ \hline
Generation by Interpolation & GPT-2 model for sentence generation and a RoBERTa coherence ranker & End sentence & Generate a specific ending \\ \hline
Plot Machines & end-to-end trainable transformer built on top of the GPT with memory representation & Outline & Generate with a storyline \\ \hline
\end{tabular}%
} \\
\end{table}
\end{appendices}

\end{document}